\documentclass[letterpaper, 10 pt, conference]{ieeeconf}

\IEEEoverridecommandlockouts
\usepackage{todonotes}
\usepackage{graphicx}
\usepackage{amsmath}
\usepackage{amssymb}
\usepackage{mathtools}
\usepackage{bm}
 \usepackage{algorithm}
\usepackage{algorithmic}
\usepackage{booktabs}
\usepackage{arydshln}
\usepackage{scalefnt}

\newcommand{\onum}[1]{\overline{{#1}}} 	
\newcommand{\unum}[1]{\underline{{#1}}}  	

\title{\LARGE \bf On-line force capability evaluation based on efficient polytope vertex search}

\author{Antun Skuric$^{1}$, Vincent Padois$^{1}$, David Daney$^{1}$ 
\thanks{*This work is supported by BPI France through the LICHIE project in collaboration with Airbus. The code of the proposed algorithm is publicly available at \text{https://gitlab.inria.fr/askuric/polytope\_vertex\_search}}
\thanks{$^{1}$  Auctus,  Inria  /  IMS  (Univ.  Bordeaux /  Bordeaux  INP  /  CNRS  UMR  5218),  33405  Talence,  France {\tt\small firstname.lastname@inria.fr}}%
}

\begin{document}
\scalefont{0.965}

\maketitle
\thispagestyle{empty}
\pagestyle{empty}

\begin{abstract}
Ellipsoid-based manipulability measures are often used to characterize the force/velocity task-space capabilities of robots. While computationally simple, this approach largely approximate and underestimate the true capabilities. Force/velocity polytopes appear to be a more appropriate representation to characterize robot's task-space capabilities. However, due to the computational complexity of the associated vertex search problem, the polytope approach is mostly restricted to  offline use, \textit{e.g.} as a tool aiding robot mechanical design, robot placement in work-space and offline trajectory planning. In this paper, a novel on-line polytope vertex search algorithm is proposed. It exploits the parallelotop geometry of actuator constraints. The proposed algorithm significantly reduces the complexity and computation time of the vertex search problem in comparison to commonly used algorithms. In order to highlight the on-line capability of the proposed algorithm and its potential for robot control, a challenging experiment with two collaborating \textit{Franka Emika Panda} robots, carrying a load of 12 kilograms, is proposed. In this experiment, the load distribution is adapted on-line, as a function of the configuration dependant task-space force capability of each robot, in order to avoid, as much as possible, the saturation of their capacity.
\end{abstract}

\section{Introduction}

Robotics manipulators and their environments are traditionally optimised for a set of specific tasks and their efficiency is based on long-term task execution. Recently, with the introduction of collaborative robots in human environments, robots need to be adaptable to unexpected events, flexible in both task definition and execution and provide high degree of safety for all humans involved. Therefore evaluating robot capabilities and optimising their performance in advance, with tools designed for more traditional robots, is no longer a suitable approach. New on-line capable and accurate evaluation techniques are needed to account for constantly changing environments, flexible tasks and interaction requirements of collaborative robots.

There are several types of metrics developed to characterise robot's kinematic, kineto-static and dynamic capabilities. The characteristics of robots are related to their kinematics, mechanical design and actuation/joint limits. Kineto-static capabilities of robots characterize the ranges of achievable twists and wrenches in arbitrary directions while in \textit{static conservative conditions}\cite{angeles_design_2016}. Dynamic capabilities characterise ranges of achievable accelerations and link elastic deformations.

Arguably the most widely used metrics are kineto-static capacity metrics based on \textit{dexterity indices}, expressing robot's ability to move and apply forces and torques in arbitrary directions with equal ease \cite{angeles_design_2016}. The widely used implementation of this metric are velocity and force manipulability ellipsoids, introduced by Yoshikawa \cite{yoshikawa_manipulability_1985}.
On the one hand, velocity manipulability ellipsoids are often used as a performance measure to optimise robot trajectories and avoid unwanted configurations. On the other hand, force manipulability ellipsoid characterises the forces the robot can apply or resist based on its kinematics. It is used during the design stage of robotic manipulators \cite{angeles_design_2016} as well as for off-line trajectory planning \cite{kuffner_motion_2016}.

\begin{figure}[!t]
    \centering
    \includegraphics[width=\linewidth]{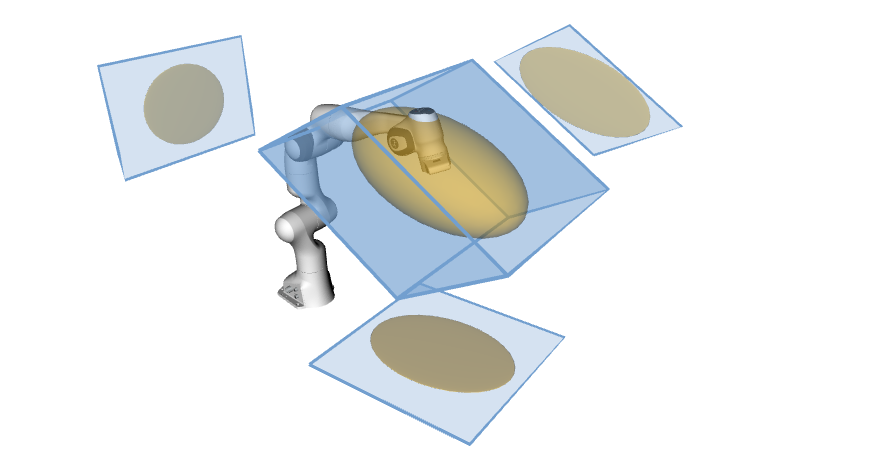}
    \caption{2D and 3D force polytopes and their ellipsoid counterparts for a 7 degrees of freedom (DoF) \textit{Franka Emika Panda} robot. Both polytopes and ellipsoids are calculated separately for the 3D and for each of the 2D reduced task-space cases. Both polytopes and ellipsoids take in consideration the true joint torque limits provided by the manufacturer. The underestimation of the true force capabilities of the robot by ellipsoids appears clearly.}

    \label{fig:polytope_ellipsoid}
    \vspace*{-0.3cm}
\end{figure}

Even though manipulability ellipsoids are widely used (mostly because of their calculation efficiency), they are just an approximation and can lead to a drastic underestimation of the real kineto-static capacities of robots \cite{merlet_jacobian_2006}. In the context of collaborative robotics, the margin between the actuation capabilities of the robot and the requirement of the task is largely reduced with respect to more classical, oversized, industrial robots. Thus, any over/under-estimation of the true capabilities of the system has strong negative implications either in terms or safety or in terms of efficiency.

Chiacchio et al.\cite{chiacchio_evaluation_1996} demonstrated that the exact wrench and twist capacities have the form of polytopes. Polytopes do not underestimate the robot capacity \cite{finotello_computation_1998}, can easily integrate variable and non-symmetric limits, dynamics and external forces \cite{merlet_jacobian_2006}. Additionally, in the context of collaborative robotics, when evaluating capacities of multiple robots, sum and intersection of ellipsoids becomes a complex problem to solve and interpret \cite{lee_dual_1989}\cite{chiacchio_task_1991}\cite{julio_frantz_analysis_2015} whereas these operations are well-defined for polytopes. Figure~\ref{fig:polytope_ellipsoid} illustrates the difference of force capability estimation between polytope and ellipsoid characterization.
However, evaluating twist and wrench polytopes relies on a \textit{vertex enumeration} problem \cite{avis_pivoting_nodate} which is a computationally intensive task and is the main obstacle for wider uses of polytopes instead of ellipsoids in practice.

In this paper, a brief overview of polytope-based task-space capability characterization is given in section~\ref{sec:section_rappel}. Then, a new on-line capable polytope vertex finding algorithm for efficiently calculating twist and wrench polytopes is proposed and described in section~\ref{sec:algorithm}. The new algorithm is capable of calculating polytope vertices for 6DoF and 7DoF robots under 3ms in Python as illustrated in section~\ref{sec:results}. This opens many doors in terms of on-line usage of such capability indicators. To demonstrate the benefits of our method for on-line robot control, an experimental setup with two collaborating \textit{Franka Emika Panda} robots is proposed. It is shown that by leveraging the real carrying capacity on-line measure of the robots, it is possible to design a control strategy allowing to carry, a voluntarily exaggerated, mass of 12 kilograms.  Discussions on the potential applications of the results of this work in collaborative robotics are then proposed in section~\ref{sec:discussion}.

\section{Task-space twist and wrench feasibility polytopes basics}
\label{sec:section_rappel}
For a serial robotic manipulator with $n$ degrees of freedom, with joint space generalized coordinates $\bm{q} \in \mathbb{R}^{n}$ and with Jacobian matrix $J(\bm{q})\in \mathbb{R}^{m \times n}$ (with $m=6$ in the 3D case and $m=3$ in the planar one), the task space Cartesian twist  $\bm{v}$ can be calculated from generalized velocities $\bm{\dot{q}}$ using
\begin{equation}
    \bm{v} = J(\bm{q})\bm{\dot{q}}
    \label{eq:velocity}
\end{equation}
The dual relation relaying the Cartesian wrench $\bm{f} \in \mathbb{R}^m$ with the generalized forces $\bm{\tau} \in \mathbb{R}^n$ is given by
\begin{equation}
    J(\bm{q})^T \bm{f} + \bm{\tau}_0 = \bm{\tau}
    \label{eq:force}
\end{equation}
$\bm{\tau}_0$ is a generalized force that does not ``produce'' any task space wrench, \textit{i.e.} $\bm{\tau}_0$ belongs\footnote{We recall here that $\mathcal{K}er(J(\bm{q})) = \mathcal{I}m(J(\bm{q})^T)^\bot$.} to the kernel $\mathcal{K}er(J(\bm{q}))$ of $J(\bm{q})$. Conversely, $(\bm{\tau}-\bm{\tau}_0)$ lies in the image  $\mathcal{I}m(J(\bm{q})^T)$ of  $J(\bm{q})^T$, \textit{i.e.} ``produces'' task space wrenches.

While equations (\ref{eq:velocity}) and (\ref{eq:force}) describe the kineto-static behaviour of a robot, generalized forces $\bm{\tau}$ and velocities $\bm{\dot{q}}$ are constrained due to the physical limits of the robot construction and actuators
\begin{subequations}
\begin{align}
    \unum{\dot{q}}_{i} \leq & \dot{q}_i \leq \onum{\dot{q}}_{i} \label{eq:constraints_v}\\
    \unum{\tau}_{i} \leq &\tau_i \leq \onum{\tau}_{i}  \label{eq:constraints_f}
\end{align}
\label{eq:constraints}
\end{subequations}
Given these constraints, the feasible twist polytope defined by equation (\ref{eq:velocity}) can be written
\begin{equation}
\begin{split}
    \mathcal{P}_v = & \left\{ \bm{v} \in \mathbb{R}^{m} \,\big| \, \bm{\dot{q}} \in [ \bm{\unum{\dot{q}}}, \bm{\onum{\dot{q}}} ],~\bm{v} = J(\bm{q})\bm{\dot{q}} \right\}
    \label{eq:polytope_v}
\end{split}
\end{equation}

Similarly, the feasible wrench polytope is defined as\footnote{The dependence of $J$ to $\bm{q}$ is dropped here and when further needed for the sake of clarity.}

\begin{equation}
\begin{split}
   \mathcal{P}_f = & \left\{ \bm{f} \in \mathbb{R}^{m}\,\big| \bm{\tau} \in \left\{[\bm{\unum{\tau}},\bm{\onum{\tau}}] \cap \mathcal{I}m(J^T)\right\},J^{T}\bm{f}=\bm{\tau} \right\}
 \label{eq:polytope_f}
\end{split}
\end{equation}


\begin{figure}[!t]
    \centering
    \includegraphics[width=\linewidth]{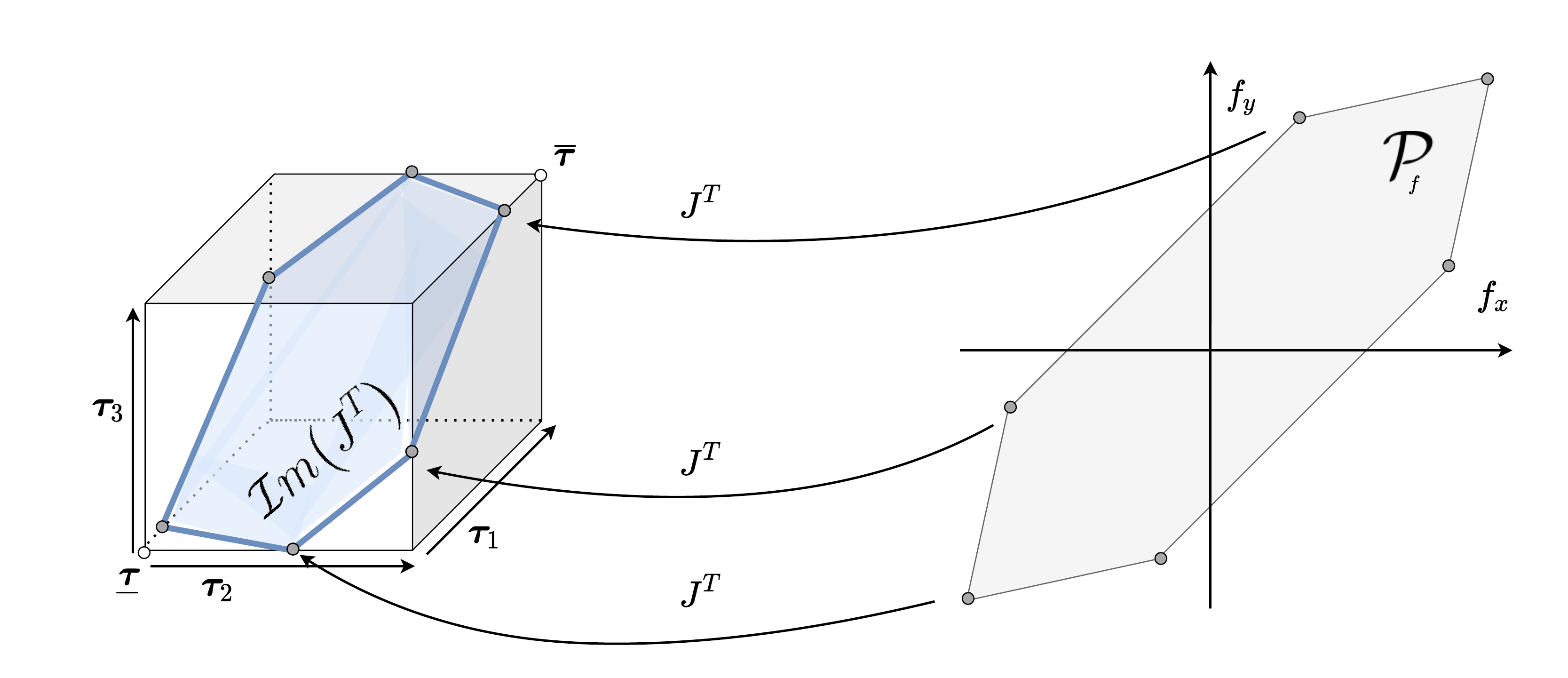}
    \caption{Example representation of force polytope $\mathcal{P}_f$ vertices in joint and task-space for a redundant 3DoF planar robot: $n$=$3$, $m$=$2$. }
    \label{fig:polytope_search}
\end{figure}

Equations (\ref{eq:polytope_v}) and (\ref{eq:polytope_f}) provide compact yet implicit representations of the feasible twist and wrench polytope of a robot. The characterization of these capabilities require an explicit definition of these polytopes, \textit{i.e.} a computation of their vertices. In the case of $\mathcal{P}_v$, computing the vertices boils down to determining the convex-hull of the image of the vertices of the constraints polytope defined by equation~(\ref{eq:constraints_v}) through the linear mapping $J(\bm{q})$. The case of $\mathcal{P}_f$ yields an extra difficulty as one first needs to determine the intersection between the constraints polytope defined by equation~(\ref{eq:constraints_f}) and $\mathcal{I}m(J(\bm{q})^T)$.

Over the years many vertex search (enumeration) methods have been proposed \cite{avis_comparative_2015}. Most of the approaches in literature are optimised to solve high dimensional problems and provide an abstraction from the actual system that is being analysed. One of the most commonly used method for vertex enumeration is the double description method \cite{avis_pivoting_nodate} which formulates the vertex search problem as transformation of an half-space representation $\mathcal{H}-rep$ to a vertex representation $\mathcal{V}-rep$. This formulation generalizes well in between different high dimensional problems but lacks the flexibility to incorporate information about the physical system being analysed.

Therefore various methods have been introduced in order to leverage specific problem formulation and improve the computational performance of vertex search. The scaling factor method has been developed for planar parallel robots \cite{nokleby_force_2005} and later adapted to planar serial manipulators \cite{julio_frantz_analysis_2015}. This method reduces the search space with one scaling factor to an exhaustive search in the scalar space. A very efficient algorithm for planar robots has also been proposed by Gouttefarde et al. \cite{gouttefarde_versatile_2015} which navigates the polytope boundaries in search for extremities. Both the scaling factor and boundaries navigation methods are designed for planar vertex search and do not scale to three dimensional world.

Chiacchio et al. in their original paper \cite{chiacchio_evaluation_1996} propose an algorithm for finding the vertices of task-space polytopes by introducing slack variables and performing an exhaustive search through the joint space limits. This algorithm, although significantly improving the evaluation is still too complex for on-line execution. This algorithm has been improved by Sasaki et al. \cite{sasaki_vertex_nodate} where the computational complexity has been significantly reduced by introducing the geometrical representation of actuator constraints as an $n$-dimensional parallelotop.

The next section introduces additional improvements of the algorithms proposed in \cite{chiacchio_evaluation_1996} and \cite{sasaki_vertex_nodate}. The resulting algorithm significantly reduces the computation time and complexity of the vertex enumeration problem for task-space polytopes. The proposed description is restricted to the most complex case of $\mathcal{P}_f$ but the method is also computationally beneficial in the simpler case of computing $\mathcal{P}_v$. Also in order to ease illustrations, the considered task-space wrench is limited to its force component without any loss of generality.

\section{Vertex finding algorithm formulation}
\label{sec:algorithm}

For an $n$DoF serial manipulator, the feasible space of joint torques $\bm{\tau}$ forms an $n$-dimensional parallelotop with $n$ pairs of parallel faces defined by the joint limits (\ref{eq:constraints_f}). The image of the Jacobian transpose matrix $\mathcal{I}m(J(\bm{q})^T)$ is a $r$-dimensional subspace of the parallelotop (where $r$ is the rank of $J(\bm{q})^T$ and $r$=$m$ for non-singular configurations\footnote{In the remainder of this paper, the robot is, without loss of generality, assumed to be in a non singular configuration.}), which corresponds to set of torques yielding non zero task space wrenches.
Sasaki et al. \cite{sasaki_vertex_nodate} have shown that the extreme values of joint torques $\bm{\tau_{vert}} \in \mathcal{I}m(J(\bm{q})^T)$ belong to the $n$-$m$ dimensional faces of the constraint parallelotop.  Furthermore, there is a one-to-one correspondence between $\bm{\tau_{vert}}$ torque vertices and the $\bm{f}_{vert}$ vertices of polytope $\mathcal{P}_f$ can be computed as
\begin{equation}
    \bm{f}_{vert} = J(\bm{q})^{T+}\bm{\tau}_{vert} \label{eq:compute_vert}
\end{equation}
where $J(\bm{q})^{T+}$ is the pseudo-inverse of $J(\bm{q})^{T}$ which provides the unique and exact solution of the inversion problem in this specific case.

Figure \ref{fig:polytope_search} shows the example of a 3DoF planar robot ($n$=$3$, $m$=$2$) for which the joint torque space is a cube and $\mathcal{I}m(J(\bm{q})^T)$ is a plane. The extreme values of feasible joint torques are found on its $n-m=1$ dimensional faces, \textit{i.e} its edges. For the example shown in figure \ref{fig:intersection_example}, $n$=$3$ and $m$=$1$ so the vertices belong to the $n-m=2$ dimensional faces of the cube, \textit{i.e.} its 2D sides.

Therefore, state-of-the-art algorithms such as \cite{chiacchio_evaluation_1996} and \cite{sasaki_vertex_nodate} propose an exhaustive search over all the $n$-$m$ dimensional parallelotop faces to find the extreme values of joint torques $\bm{\tau_{vert}}$ and consequently the vertices of force polytope $\mathcal{P}_f$. In this paper, a new representation of joint torque vector $\bm{\tau}$ is proposed enabling an efficient navigation of the $n-m$ dimensional parallelotop faces, reducing the complexity of the exhaustive search.

\subsection{Proposed algorithm}

Consider a joint torque vector $\bm{\tau}$ meeting constraints~\ref{eq:constraints_f}. It can be defined as
\begin{equation}
    \bm{\tau} = \unum{\bm{\tau}} + \alpha_1 \bm{\tau}_1+ \alpha_2 \bm{\tau}_2 + ... + \alpha_n \bm{\tau}_n
    \label{eq:torque_new}
\end{equation}
where $\alpha_i \in [0,1]$ are scalars weights and vectors $\bm{\tau}_i$ are orthogonal vectors in joint space aligned with $i$-th axis of the parallelotop, defined as $\bm{\tau}_i = \begin{bmatrix} 0~\ldots~\onum{\tau}_{i} - \unum{\tau}_{i}~\ldots~0 \end{bmatrix}^T$.

Since the vertices $\bm{\tau_{vert}} \in \mathcal{I}m(J(\bm{q})^T)$ belong to the $n$-$m$ dimensional faces of the joint torque parallelotop, representation (\ref{eq:torque_new}) provides an elegant way to navigate them, by fixing $m$ out of $n$ scalars $\alpha_i$ either to $0$ or to $1$.

\begin{figure}[!t]
    \centering
    \hspace*{-0.5cm}
    \includegraphics[width=1.1\linewidth]{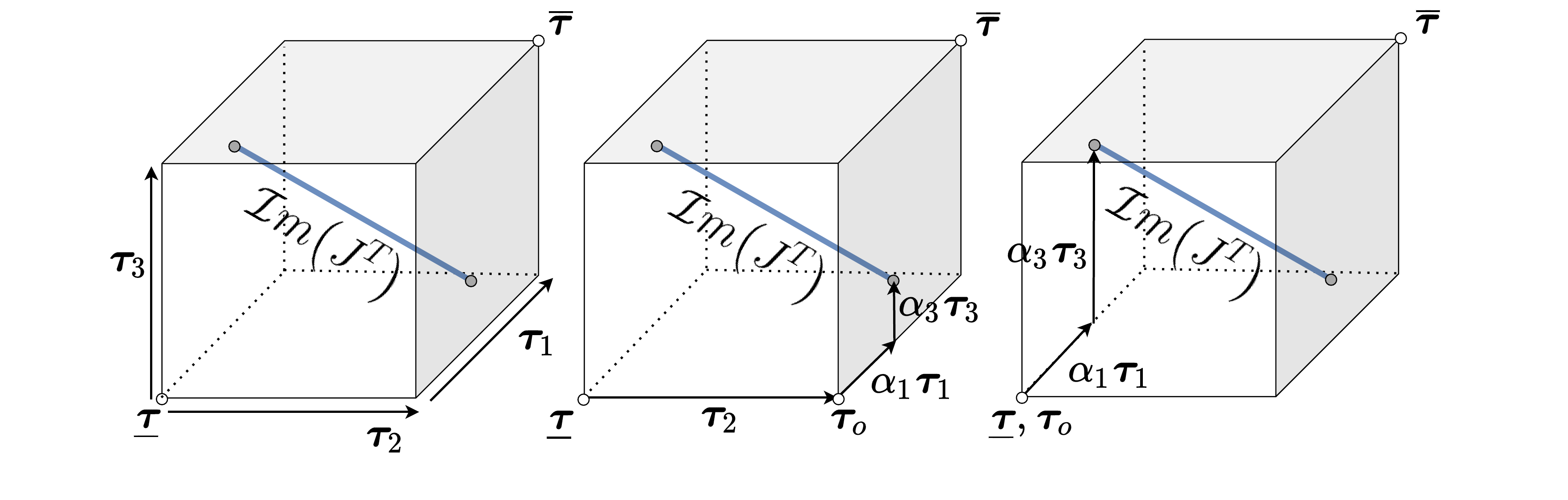}
    \caption{Example of joint space interpretation of the vertex search algorithm solution for $n$=$3$ and $m$=$1$. }
    \label{fig:intersection_example}
    \vspace*{-0.3cm}
\end{figure}

An $n$-dimensional parallelotop has $\big(\begin{smallmatrix}n\\m\end{smallmatrix}\big)$ sets of $2^m$ parallel $n$-$m$ dimensional faces. Each set of parallel faces can be represented with the linear combination of the same set of $n$-$m$ base vectors $\bm{\tau}_i$ or in other words with a set of $n$-$m$ scalars $\alpha_i$. The remaining $m$ scalars are set either to 0 or 1 and define the origin $\bm{\tau}_o$ of  each one of the $2^m$ faces from the set
\begin{equation}
    \bm{\tau}_o = \unum{\bm{\tau}} + \alpha_1 \bm{\tau}_1+ ... + \alpha_m \bm{\tau}_m
\end{equation}
Therefore each $n$-$m$ dimensional face can be reached by partitioning $\bm{\alpha} = \left[\bm{\alpha}_1^T ~\bm{\alpha}_2^T\right]^T$, $\bm{\alpha}_1$ containing $m$ scalars $\alpha_i$ fixed to $0$ or $1$ and $\bm{\alpha}_2$ containing the remaining $n$-$m$ scalars which define the linear combination of $n$-$m$ base vectors $\bm{\tau}_i$.

Figure \ref{fig:intersection_example} illustrates the case where $\mathcal{I}m(J^T)$ is a line in joint space. In this case the intersection space is $n$-$m$=$2$ dimensional. In this case, two scalars $\alpha_1$ and $\alpha_3$ are enough to characterise both vertices of interest with $\alpha_2$ fixed to $0$ or $1$.

In order to find the vertices of the force polytope $\mathcal{P}_f$, equations (\ref{eq:force}) and (\ref{eq:torque_new}) are combined to restrict the search to the set of torques $\bm{\tau} \in \left\{[\bm{\unum{\tau}},\bm{\onum{\tau}}] \cap \mathcal{I}m(J^T)\right\}$
\begin{equation}
    J(\bm{q})^T \bm{f}_{vert} = \unum{\bm{\tau}} + \alpha_1 \bm{\tau}_1+ \alpha_2 \bm{\tau}_2 + ... + \alpha_n \bm{\tau}_n \label{eq:inter_im_bounds_torque}
\end{equation}
For each $\left(\begin{smallmatrix}n\\m\end{smallmatrix}\right)$ combination of the $m$ scalars, $2^m$ values of $\bm{\tau}_o$ can be calculated. For each possible value of $\bm{\tau}_o$, a linear system deriving from equation~(\ref{eq:inter_im_bounds_torque}) can be solved to find the $n$-$m$ scalars $\bm{\alpha}_{2}$ and the corresponding force polytope vertex $\bm{f}_{vert}$
\begin{equation}
    \underbrace{\begin{bmatrix}J(\bm{q})^T \,-\bm{\tau}_{m+1} \, \dots \, -\bm{\tau}_{n} \end{bmatrix}}_{Z_{n\times n}} \begin{bmatrix}\bm{f}_{vert}\\ \alpha_{m+1} \\ \vdots\\\alpha_{n} \end{bmatrix} = \bm{\tau}_o
    \label{eq:linear_system_full}
\end{equation}
If $Z$ is invertible and if $\bm{\alpha}_2 \in [\bm{0},\bm{1}]$, $\bm{f}_{vert}$ is a vertex of the polytope $\mathcal{P}_f$
\begin{equation}
   \begin{bmatrix}\bm{f}_{vert}\\ \bm{\alpha}_{2} \end{bmatrix} = Z^{-1}\bm{\tau}_L, \quad  \bm{\alpha}_{2} \in [\bm{0},\bm{1}]
    \label{eq:linear_system_full}
\end{equation}

Overall, in order to navigate all possible combinations of the $n$-$m$ dimensional faces which may contain vertices of the force polytope $\mathcal{P}_f$, $\big(\begin{smallmatrix}n\\m\end{smallmatrix}\big)$=$\frac{n!}{m!(n-m)!}$ inversions of $Z$ and  $2^m \frac{n!}{m!(n-m)!}$ checks that $\bm{\alpha}_2 \in [\bm{0},\bm{1}]$ have to be performed.

In the case depicted in figure \ref{fig:intersection_example} ($n$=$3$, $m$=$1$), $Z$ is inverted only $\big(\begin{smallmatrix}3\\1\end{smallmatrix}\big)$=$3$ times and conditions are evaluated 6 times, which exactly corresponds to the number of faces of the parallelotop. Taking the example of the 3DoF planar robot ($n$=$3$, $m$=$2$) the number of $Z$ inversions is 3 and the number of condition evaluation is 12, which exactly corresponds to the number of edges of the parallelotop. One of the nice features of this approach is that it does not require an explicit inversion of $J(\bm{q})^T$ as in equation~(\ref{eq:compute_vert}) to actually obtain the set of vertices composing $\mathcal{P}_f$.

\subsection{Improvement using the SVD}

In order to further improve the efficiency of the proposed algorithm the Singular Value Decomposition (SVD) \cite{klema_singular_1980} of $J(\bm{q})$ is used
\begin{equation}
     J(\bm{q}) =  U  \underbrace{\begin{bmatrix}S & O_{m\times (n-m)}\end{bmatrix}}_{\Sigma}\underbrace{\begin{bmatrix}V_1^T \\ V_2^T \end{bmatrix}}_{V^T}
\end{equation}
where $S= diag( \sigma_1 \dots \sigma_m)$ is a diagonal matrix containing the $m$ singular values of $J(\bm{q})$. $V_1^T \in \mathbb{R}^{m \times n}$ projects the space of generalized velocities onto the preimage of $\mathcal{I}m(J(\bm{q}))$ and $V_2^T \in \mathbb{R}^{n-m \times n}$ is a basis of $\mathcal{K}er(J(\bm{q}))$. $V_1^T$ and $V_2^T$ are orthogonal vector subspaces yielding $V_2^T V_1 = O$ and thus
\begin{equation}
    V_2^T J(\bm{q})^T\bm{f} = \bm{0}
\end{equation}
This allows to reduce (\ref{eq:linear_system_full}) to
\begin{equation}
    \underbrace{V_2^T\begin{bmatrix}-\bm{\tau}_{m+1} \dots -\bm{\tau}_{n} \end{bmatrix}}_{T_{(n-m)\times (n-m)}} \bm{\alpha}_2 = V_2^T\bm{\tau}_o
    \label{eq:linear_system_svd}
\end{equation}
If $T$ is invertible, $\bm{\alpha}_2$ can be computed by
\begin{equation}
\bm{\alpha}_2 = T^{-1}V_2^T\bm{\tau}_o
\end{equation}
When the computed $\bm{\alpha}_2 \in [\bm{0},\bm{1}]$,  the corresponding force polytope vertex can be calculated as
\begin{equation}
    \bm{f}_{vert} = J(\bm{q})^{T+} \big( \unum{\bm{\tau}} + \alpha_1\bm{\tau}_{1} +\dots+\alpha_n\bm{\tau}_{n}\big)
\end{equation}

This new approach requires the calculation of the SVD and the Jacobian matrix pseudo-inverse $J(\bm{q})^{T+}$. Nevertheless, since the Jacobian transpose pseudo-inverse can be efficiently calculated from the SVD as $J(\bm{q})^{T+} = U\Sigma^{T+}V^T$
and since both the SVD and pseudo-inverse are calculated once and of all per algorithm run, the computation efficiency is greatly improved due to the matrix dimension reduction from $n$ to $(n-m)$ when inverting $T$ instead of $Z$.

\subsection{Matrix inverse condition}

Due to the constrained nature of $\bm{\alpha}_2$, it is possible to efficiently calculate some bounds ${\bm{t}_{ub}}$ and $\bm{t}_{lb}$ such that $T\bm{\alpha}  \in [\bm{t}_{lb}, \bm{t}_{ub}]$.
These bounds are found by row-wise summing only positive and only negative elements $t_{ij}$ of  $T$
\begin{equation}
t_{i,ul} = \sum_j \max(t_{ij}, 0), \qquad
t_{i,lb} = \sum_j \min(t_{ij}, 0)
\label{eq:bounds_condition}
\end{equation}

This creates a bounding box around the space defined with the vector $T\bm{\alpha}_2$. Therefore, one can conclude that if all the $2^m$ combinations of $\bm{\tau}_o \notin [\bm{t}_{lb}, \bm{t}_{ub}] $, the system (\ref{eq:linear_system_svd}) cannot have a solution which satisfies $\bm{\alpha}_2 \in [\bm{0},\bm{1}]$, and there is no need to invert $T$ to figure it out.

\subsection{Extension to residual polytopes}

When evaluating the capability of a robot, it is a common practice to take into consideration the joint torques necessary to compensate for gravity $\bm{\tau}_g = \bm{g}(\bm{q})$ \cite{wei_output_nodate}
\begin{equation}
    J(\bm{q})^T \bm{f} = \bm{\tau} - \bm{\tau}_g
\end{equation}
Furthermore, the same can be done to include the effects of the robot dynamics. For a fixed-base, serial manipulator the dynamics can be written as
\begin{equation}
    M(\bm{q})\ddot{\bm{q}} + C(\dot{\bm{q}},\bm{q})\dot{\bm{q}} + \bm{g}(\bm{q}) = \bm{\tau} - J(\bm{q})^T \bm{f}
    \label{eq:full_dynamics}
\end{equation}
where $M$ and $C$ are respectively the mass and Coriolis-Centrifugal matrices. The residual joint torque vector $\bm{\tau}$ can then be expressed as
\begin{equation}
    J(\bm{q})^T \bm{f} = \bm{\tau} - \bm{\tau}_g  - \big(\underbrace{M(\bm{q})\ddot{\bm{q}} + C(\dot{\bm{q}},\bm{q})\dot{\bm{q}}) \big)}_{\bm{\tau}_{d}}
    \label{eq:force_dynamics}
\end{equation}

Finally, in many cases it useful to evaluate the residual force ability of a robot already applying a certain nominal wrenches $\bm{f}_{n}$ which imply to generate joint torques $\bm{\tau}_{n}$
\begin{equation}
\bm{\tau}_{n} =  J(\bm{q})^T \bm{f}_{n}
\end{equation}

The \textit{residual force polytope} \cite{ferrolho_residual_2020} corresponding to the  feasible space of Cartesian wrenches $\bm{f}$ that can be applied and rejected by the robot end-effector, taking into consideration the torque necessary to overcome gravity $\bm{\tau}_{g}$, robot dynamics $\bm{\tau}_{d}$ and any nominal joint torque $\bm{\tau}_{n}$, can be computed using algorithm~\ref{alg:main_algo} and the updated joint constraints
\begin{equation}
\begin{split}
    \unum{\bm{\tau}}' &= \unum{\bm{\tau}} - \bm{\tau}_g - \bm{\tau}_{d} - \bm{\tau}_{n}  \\
    \onum{\bm{\tau}}' &= \onum{\bm{\tau}}  - \bm{\tau}_g - \bm{\tau}_{d} - \bm{\tau}_{n}
    \label{eq:residual_constraints}
\end{split}
\end{equation}



\begin{algorithm}[!t]
\caption{New vertex search algorithm pseudo-code}
\begin{algorithmic}
\REQUIRE $J$, $\onum{\bm{\tau}}$, $\unum{\bm{\tau}}$ (Eq. \ref{eq:constraints_f} or Eq. \ref{eq:residual_constraints})
\STATE $U, \Sigma, V^T \leftarrow svd(J)$
\STATE $J^{T+} = U\Sigma^{T+}V^T$
\STATE $V_1,\, V_2  \leftarrow V $
\STATE calculate $n$ base vectors $\bm{\tau}_1 \dotsc \bm{\tau}_n$ (Eq. \ref{eq:torque_new})

\FORALL{  $\big(\begin{smallmatrix}n\\m\end{smallmatrix}\big)$ combinations of $m$ fixed $\alpha_i$ }
\STATE construct $T$ matrix (Eq. \ref{eq:linear_system_svd})
\STATE find bounds $\bm{t}_{lb},\bm{t}_{ub}$ (Eq. \ref{eq:bounds_condition})

\FORALL{   $2^m$ vectors $\bm{\tau}_o$ }
\IF{$ V_2^T\bm{\tau}_o \in [\bm{t}_{lb},\bm{t}_{ub}] $}
\STATE $\bm{\alpha}_2 = T^{-1}V_2^T\bm{\tau}_o $
\IF{$ \bm{\alpha}_2 \in [0,1] $}
\STATE $\bm{\tau}_{vert} = \unum{\bm{\tau}} + \alpha_1{\bm{\tau}_{1}} +\dots+\alpha_n{\bm{\tau}_{n}}$
\STATE $\bm{f}_{vert} = J^{T+}\bm{\tau}_{vert}$
\ENDIF
\ENDIF
\ENDFOR
\ENDFOR

\end{algorithmic}
\label{alg:main_algo}
\end{algorithm}

\begin{figure*}[!t]
    \centering
    \includegraphics[width=\linewidth]{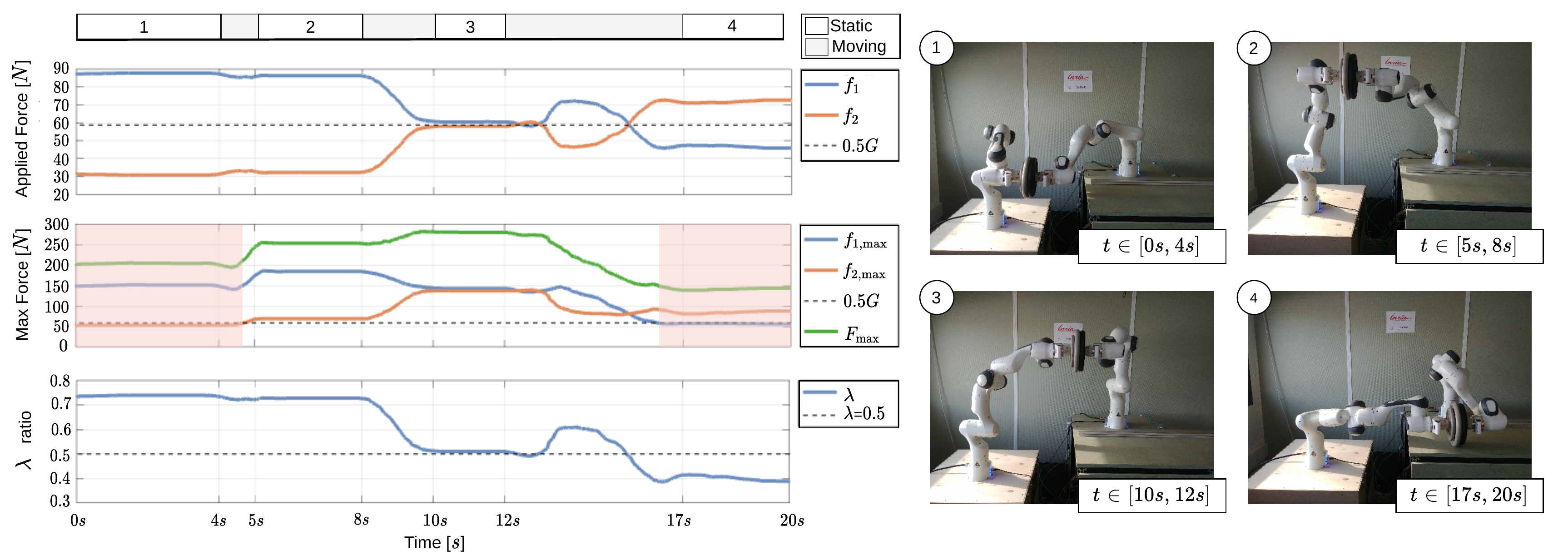}
    \caption{Images 1 to 4 on the right show the sequence of poses the manipulators is placed by a human operator during the course of the experiment. On left side, on the top is the graph showing the time evolution of applied manipulator forces $f_1,f_2$. The middle graph provides the evolution of the maximal applicable forces $f_{i,max}$ of each manipulator separately and their joint capacity $F_{max}$, based on the proposed polytope evaluation algorithm. In this graph,  the time periods when the constant strategy $\lambda$=$0.5$ (dashed line on all graphs) would not be feasible for at least one of the two robot are indicated with red background. The bottom graph shows the  evolution in time of the control parameter $\lambda$ for the proposed adaptive strategy.}
    \label{fig:dual_manip}
    \vspace*{-0.3cm}
\end{figure*}

\section{Experiments and results}
\label{sec:results}
In this section, the complexity of the proposed algorithm is evaluated for different robots and compared to state-of-the-art algorithms. Furthermore, a real-time control strategy for dual arm collaborative object handling using the proposed on-line polytope estimation is introduced and experimentally validated  in  section \ref{sec:colaborative_handling}.

\subsection{Complexity analysis}\label{sec:complexity}
To demonstrate the efficiency of the proposed polytope vertex search algorithm, it is compared to the polytope vertex search algorithm introduced by Chiacchio et al. \cite{chiacchio_evaluation_1996}. Furthermore the comparison is extended to the algorithm proposed by Sasaki et al. \cite{sasaki_vertex_nodate} which is, to our knowledge, the only algorithm exploiting the geometric structure of the problem.

The three algorithms have been tested to find vertices of the force polytope $\mathcal{P}_f$ for three different robots: a 4R planar robot ($n$=$4$,$m$=$2$), the \textit{Universal Robots} UR5 6DoF robot ($n$=$6$,$m$=$3$) and the \textit{Franka Emika Panda} 7DoF robot ($n$=$7$,$m$=$3$). The results are averaged over 1000 randomly selected robot configurations. All algorithms have been implemented in the programming language Python and tested on a laptop equipped with a 1.90GHz Intel i7-8650U processor.

Table \ref{tab:complexity_results} shows the results of this complexity evaluation. The proposed vertex search algorithm drastically reduces the number of matrix inversions and thus reduces the processing time considerably: $4$-$10\times$ faster execution than Chiacchio's algorithm and on average 20\% faster than Sasaki's. Results show that even for the cases of 6DoF and 7DoF industrial robots our approach is capable of evaluating the polytope vertices under $3ms$. Such a low processing time opens numerous opportunities for the on-line use of polytope based capacity evaluation especially in the area of robot control.

\begin{table}[!b]
    \centering
    \caption{Complexity and execution time comparison for three different vertex search algorithms.}
    {\scalefont{0.99}
    \begin{tabular}{lccc}
       Robot & \textbf{Chiacchio}\cite{chiacchio_evaluation_1996} & \textbf{Sasaki} \cite{sasaki_vertex_nodate}  &  \textbf{Proposed} \\
       \hline \hline
       $(n,m)$& \multicolumn{1}{l}{inversions: $ \frac{2n!}{m!(2n-m)!}$} & $ \frac{n!}{m!(n-m)!} $ & $ \leq \frac{n!}{m!(n-m)!}$ \\
       & \multicolumn{1}{l}{mat. size:  $ 2n \times 2n$} & $m\times m$  & $n$-$m$ $\times$ $n$-$m$\\
       & \multicolumn{1}{l}{time[ms]: t $\pm$ sd (max)} & & \\
       \hline
       $(4,2)$ &  24 & 6 & 4.2$\pm$1.4 (6) \\
       & 8$\times$8 & 2$\times$2 & 2$\times$2 \\
       & 2.7$\pm$0.1 (3.6) & 0.93$\pm$0.1 (1.1) & 0.64$\pm$0.1 (1.5) \\
       \hline
       $(6,3)$ & 220 & 20 & 2.8$\pm$2.5 (10) \\
        & 12$\times$12 & 3$\times$3 & 3$\times$3 \\
       & 14.2$\pm$0.9 (20) & 2.1$\pm$0.2 (3.4) & 1.5$\pm$0.2 (2.7) \\
       \hline
      $(7,3)$&    364 & 35 & 7.3$\pm$4.2 (20)\\
        & 14$\times$14 & 3$\times$3 & 4$\times4$\\
       & 25$\pm$0.5 (27) & 3.5$\pm$0.1 (5.4) & 2.6$\pm$0.2 (4.3)\\

    \end{tabular}
    }
    \label{tab:complexity_results}
\end{table}

\subsection{Dual arm collaborative object carrying}\label{sec:colaborative_handling}

In this section, the real-time force capability evaluation is employed in the context of robot control. An experiment is designed using two \textit{Franka Emika Panda} robots involved in a collaborative load carrying task for an object of mass $M$. Each robot is contributing to the compensation of the overall gravity force $G=Mg$ applied at the end-effector by the carried object, with forces $f_1$ and $f_2$ respectively such that
\begin{equation}
    f_1 + f_2 = G
\end{equation}ICRA2021
Additionally, a human operator is moving the object freely through the common work-space, implicitly changing both robots positions and configurations in real-time. As a consequence, the task space trajectory and the evolution of the robot configurations are not known in advance.

The experiment has been implemented using the ROS programming environment. Both robots are torque controlled at $1 kHz$. The update of the force capability estimation are performed at $40 Hz$. All the control software is run from one computer with the 1.90GHz Intel i7-8650U processor. This experiment is illustrated in the accompanying video.

Panda robots are rated to a maximal carrying capacity of 3kg which corresponds to the absolute minimal carrying capacity of the robot evaluated in one of its near-singular configurations. It is, by definition, an underestimation of the real robot task space wrench capacity and relying on it, as it is commonly done, limits the scope of the possible tasks and applications.

The goal of this experiment, it is to demonstrate the fact that, taking into account the true force capabilities of each robot, it is possible to considerably go beyond the robots conservative rated capacity without comprising safety nor exceeding any of the actuation limits. The weight of the object chosen for the experiments is $M=12kg$, voluntarily far above the recommended joint carrying capacity $M \le 6kg$.

Evaluating the maximal applicable force in the vertical direction can be viewed as a special case of the force polytope $\mathcal{P}_f$ vertex search problem where the task-space force of interest $f$ is a scalar and $m=1$.
The proposed algorithm described in section \ref{sec:algorithm} is used to efficiently find $f_{i,max}$, the limits (vertices) of the maximal applicable force in $z$-axis direction, for each robot separately.  Their overall carrying capacity can be calculated as the sum of the two $F_{max} = f_{1,max} + f_{2,max}$.

To efficiently distribute and adapt the carried load in between robots, each robot compensate for a part of the object's weight $G$ which is proportional to its force capacity  $f_{i,max}$
\begin{equation}
   f_1 = \lambda G \quad f_2 = (1-\lambda)G,\quad \lambda = \frac{f_{1,max}}{f_{1,max} + f_{2,max}}
\end{equation}

A straightforward control strategy, requiring each robot to compensate for half of object mass $f_1=f_2=0.5G$ is taken as baseline to evaluate the performance of the new method.

Figure \ref{fig:dual_manip} (right) shows one possible manipulation trajectory, where a human operator displaces the carried object in the common workspace in four discrete locations over a 20 seconds period of time. On the left side of the figure \ref{fig:dual_manip} the evaluated maximal forces and force applied by each robot during the experiment are shown. Additionally, the $0.5G$ strategy line is shown for comparison.

The proposed control strategy is successful in ensuring  compensation of the object weight during the full length of the experiment. In the starting pose $t=0s$, robot 2 (robot on the left) is close to it's singular configuration and its load carrying capacity is close to 5kg. Controlling it to compensate for half of the weight of the object would result in a security exception and could damage the robot hardware. The same is true for pose 4 at $t=17s$. Robot 1 (robot on the right) is not be able to compensate for half of the object weight but thanks to the proposed adaptive control law, the task can successfully be achieved. This may not be the case over the entire common workspace of the two robots but this example is a good illustration of the interest of accounting for the true capabilities of the system at the control level.  An additional interesting result from the experiment is the transition between pose 3 and 4 ($t \in [12s,16s]$). Even though the object is much closer to robot 2 (right), robot 1 has a much higher carrying capacity. This illustrates the fact that force capabilities are highly nonlinear functions which are very hard to predict. This provides an additional argument in favor of on-line force capacity evaluation.

\section{Discussions}
\label{sec:discussion}
In this section, the potential of applying the on-line capacity evaluation for collaborative robots and the human-robot interaction is discussed.

\subsection{Polytopes for multi-robot systems}
\begin{figure}
    \centering
    \includegraphics[width=\linewidth]{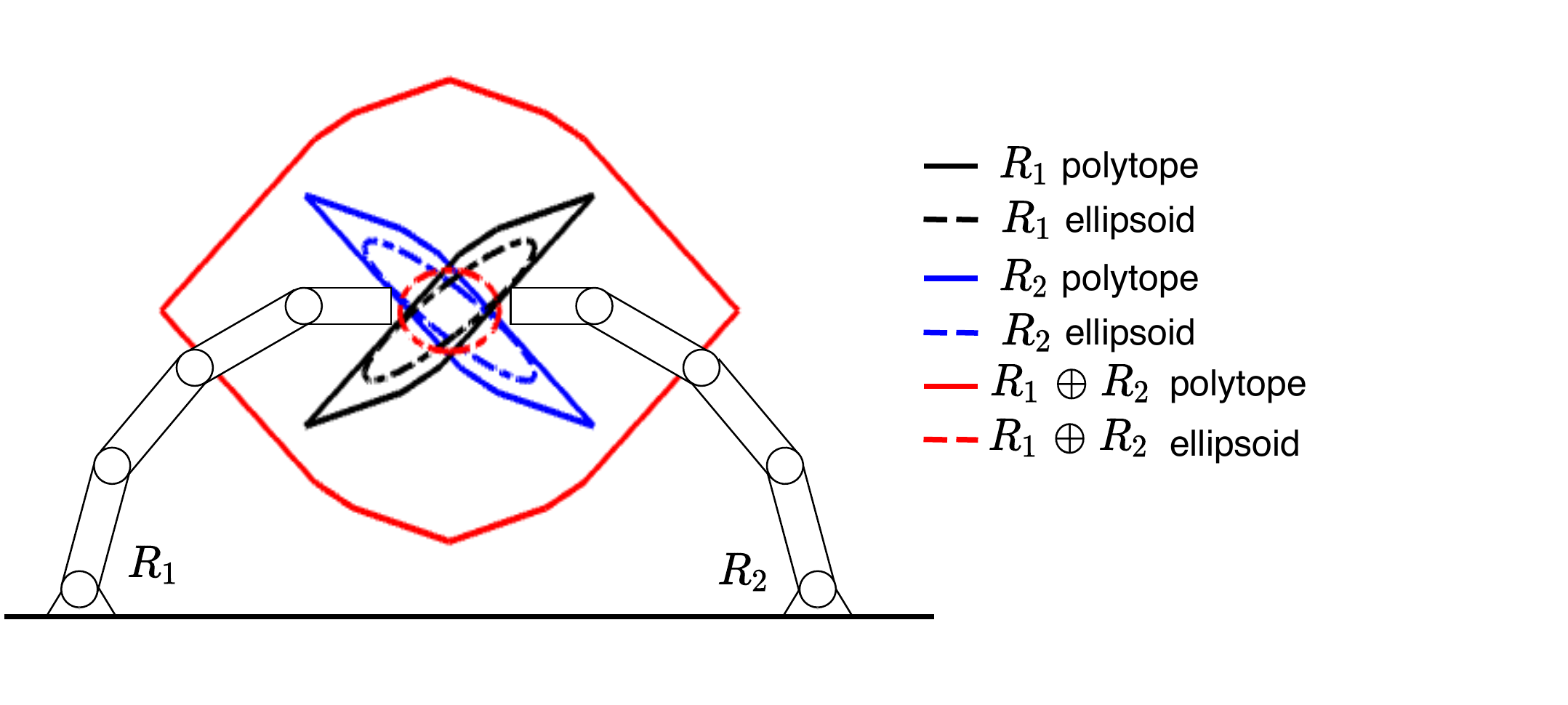}
    \caption{4DoF robots collaboration: polytopes and ellipsoids comparison}
    \label{fig:collaborative_difference}
    \vspace*{-0.3cm}
\end{figure}

Up to recently \cite{long2020constrained}, the estimation of task-space capabilities of multiple robotic manipulators was largely depending on manipulability ellipsoids \cite{chiacchio_task_1991}\cite{chiacchio_global_1991}\cite{lee_dual_1989}.

However as the number of robots grows, the ellipsoid approach, which rely on a Euclidean norm rather than an infinity norm \cite{merlet_jacobian_2006}, provides a less accurate approximation and the questions of what is the best practice to sum \cite{chiacchio_task_1991}\cite{julio_frantz_analysis_2015} or intersect \cite{sekiguchi_force_2017}\cite{lee_dual_1989} multiple ellipsoids becomes problematic.

Figure \ref{fig:collaborative_difference} shows one example of two collaborating 4DoF planar robots joint force capacity based on ellipsoids and polytopes. The ellipsoids are calculated using the approach proposed by Chiacchio \cite{chiacchio_task_1991} and the joint polytope is calculated as the Minkowski sum of the individual force polytope of each robot.
\begin{equation}
\mathcal{P}_{f_\oplus} = \mathcal{P}_{f_{R_1}} \oplus \mathcal{P}_{f_{R_2}}
\end{equation}
This example clearly demonstrates that even though ellipsoids can reasonably well provide an approximation of the force capacity of each robot separately, the same is not the case for the capacity of the dyad.

A similar conclusion can be drawn for use cases where the joint capacity of multiple robots can be represented by the intersection of their polytopes \cite{long2020constrained}, \textit{i.e} when the robots are working in opposition, both applying a similar force $\bm{f}$ in opposite directions\footnote{A typical example of this situation is the case of the frictional grasp of an object.}
\begin{equation}
\mathcal{P}_f = \mathcal{P}_{f_{R_1}}  \cap \mathcal{P}_{f_{R_2}}
\end{equation}
In that case, the polytope intersection problem can be elegantly transformed into the vertex enumeration problem for the extended system
\begin{equation}
J_\cap = [J_{R_1}\, J_{R_2}],\quad \bm{\tau} = \left[\bm{\tau}_{R_1}^T\, \bm{\tau}_{R_2}^T\right]^T, \quad J_\cap^T\bm{f} = \bm{\tau}
\end{equation}

\subsection{Polytopes for human-robot collaboration}
Several authors have studied the use of robotics performance measures such as manipulability ellipsoids and polytopes for human arms using 7DoF manipulator models \cite{rezzoug_application_2012}\cite{lazinica_higher_2010}\cite{carmichael_estimating_2013}\cite{carmichael_towards_2011} or full musculoskeletal model \cite{hernandez_toward_2015}. A promising approach for leveraging the efficient on-line polytope evaluation algorithm is to estimate the joint capacity of the robot and human and adapt the robotic assistance as a function of the evolution of human capabilities.  One of the associated challenges is to extend the proposed vertex search algorithms to actuation constraints which result from more complex actuation mechanisms such as muscles and can thus no longer be represented as parallelotops.

\section{Conclusion}

The proposed vertex search algorithm provides a computationally efficient tool for the polytope-based, online evaluation of task-space robot capabilities. As illustrated in this work, the ability to accurately evaluate on-line the capabilities of robots leverages several possibilities in modern robotic applications where oversizing robots is no longer an option.






\newpage
\bibliographystyle{ieeetr} 
\bibliography{skuric_padois_daney_submitted_nov_2020}

\end{document}